\pdfoutput=1
\documentclass{article}

\usepackage[preprint]{neurips_2021}
\usepackage{newtxtext}
\usepackage{newtxmath}
\usepackage[hidelinks]{hyperref}
\usepackage[english]{babel}
\usepackage{amsmath}
\usepackage{amsfonts}
\usepackage{bm}
\usepackage{nicefrac}
\usepackage{microtype}
\usepackage[table]{xcolor}
\usepackage{tikz}
\usepackage{pgfplots}
\usepackage{xifthen}
\usepackage{caption}
\usepackage{subcaption}
\usepackage{xspace}
\usepackage{enumitem}
\usepackage[textwidth=1.7cm,textsize=small,disable]{todonotes}
\usepackage{dcolumn}
\usepackage{array}
\usepackage{natbib}
\usepackage{booktabs}

\def\x{\vct{x}} %
\def\original{\text{R}}
\def\counterfactual{\text{C}}
\def\hybrid{\text{H}}
\def\xo{\x_\original} %
\def\xc{\x_\counterfactual} %
\def\xh{\x_{\hybrid}} %
\def\yo{y_\original} %
\def\yc{y_\counterfactual} %
\def\yh{y_{\hybrid}} %
\def\am{\vct{a}}  %
\def\mask{\vct{m}}  %

\def\pred{D} %

\def\mnist{\textsc{Mnist}\xspace}
\def\synapse{\textsc{Synapses}\xspace}
\def\disc{\textsc{Disc}\xspace}

\def\vgg{\textsc{VGG}\xspace}
\def\resnet{\textsc{ResNet}\xspace}
\def\operation#1{\texttt{#1}\xspace}

\def\vct#1{\bm{#1}}

\newcommand{\argmin}[1]{\mathop{\arg\min}_{#1}\hspace{0.5em}}

\usetikzlibrary{arrows}
\usetikzlibrary{arrows.meta}
\usetikzlibrary{backgrounds}
\usetikzlibrary{calc}
\usetikzlibrary{chains}
\usetikzlibrary{decorations.markings}
\usetikzlibrary{decorations.pathmorphing}
\usetikzlibrary{fit}
\usetikzlibrary{matrix}
\usetikzlibrary{mindmap}
\usetikzlibrary{pgfplots.colormaps}
\usetikzlibrary{positioning}
\usetikzlibrary{shadows}
\usetikzlibrary{shapes}
\usetikzlibrary{trees}
\usetikzlibrary{spy}
\usetikzlibrary{intersections}

\definecolor{funkey_bright}{HTML}{FFFFFF}
\definecolor{funkey_lightgrey}{HTML}{CCCCCC}
\definecolor{funkey_grey}{HTML}{666666}

\definecolor{funkey_color_1}{HTML}{834D9D}
\definecolor{funkey_color_2}{HTML}{F2A431}
\definecolor{funkey_color_3}{HTML}{55B849}

\definecolor{funkey_color_4}{HTML}{DB8457}%
\definecolor{funkey_color_5}{HTML}{8174B1}%
\definecolor{funkey_color_6}{HTML}{ADD8E6}%
\definecolor{funkey_color_7}{HTML}{000080}%
\definecolor{funkey_color_8}{HTML}{800020}%
\definecolor{funkey_color_9}{HTML}{228B22}%
\definecolor{funkey_color_10}{HTML}{F5C108}%
\definecolor{funkey_color_11}{HTML}{DA3074}%
\definecolor{funkey_color_12}{HTML}{CC00FF}%
\definecolor{funkey_color_13}{HTML}{FF9900}%
\definecolor{funkey_color_14}{HTML}{006666}%

\definecolor{fake_color}{HTML}{FFB31A}%
\definecolor{real_color}{HTML}{00CCFF}%

\colorlet{funkey_dark}{purple!10!black}

\colorlet{funkey_highlight}{funkey_color_2}
\colorlet{funkey_textcolor}{funkey_bright}
\colorlet{funkey_bg}{funkey_dark}
\colorlet{funkey_alt_bg}{funkey_lightgrey}

\definecolor{color_gaba}{HTML}{F2A431}
\definecolor{color_ach}{HTML}{8EA4D2}
\definecolor{color_glut}{HTML}{55B849}
\definecolor{color_ser}{HTML}{D237C5}
\definecolor{color_oct}{HTML}{834D9D}
\definecolor{color_dop}{HTML}{DE6868}

\definecolor{color_1}{HTML}{82CAFC}
\definecolor{color_2}{HTML}{C875C4}
\definecolor{color_3}{HTML}{580F41}
\definecolor{color_4}{HTML}{2E294E}
\definecolor{color_5}{HTML}{C2BBF0}

\definecolor{color_bright}{HTML}{FFFFFF}
\definecolor{color_lightgrey}{HTML}{EEEEEE}
\definecolor{color_grey}{HTML}{666666}
\definecolor{color_dark}{HTML}{3D3D35}
\colorlet{color_highlight}{color_2}

\tikzstyle{icon}=[rectangle,rounded corners,draw=funkey_lightgrey,thick]

\tikzstyle{neuron}=[circle,fill=funkey_color_2,inner sep=0.7mm]
\tikzstyle{synapse}=[->,>=latex',funkey_color_1,thick]

\tikzstyle{inset}=
  [rectangle,inner sep=0,clip]
\tikzstyle{candidate}=
  [very thick,funkey_color_1,->,shorten >=2mm]
\tikzstyle{cell}=
  [circle,inner sep=3pt,fill]

\tikzstyle{data}=
  [rectangle,rounded corners,draw,fill=funkey_lightgrey]
\tikzstyle{method}=
  [rectangle,rounded corners,fill=funkey_color_1,text=funkey_bright,text width=5mm,align=center]
\tikzstyle{arrow}=
  [->,very thick,>=latex']
\tikzstyle{process}=
  [single arrow,fill=funkey_lightgrey,inner sep=2pt,minimum height=5mm,single arrow head extend=1.5mm]
\tikzstyle{image}=
  [inner sep=0,outer sep=0,draw=funkey_lightgrey]

\tikzstyle{perceptron}=
  [circle,outer color=orange!50!white,inner color=white]
\tikzstyle{perceptron_input}=
  [circle,outer color=blue!20!white, inner color=white,anchor=base]
\tikzstyle{perceptron_output}=
  [circle,outer color=blue!20!white, inner color=white,anchor=base]
\tikzstyle{operator}=
  [circle,draw=lightgray,fill=white,inner sep=1pt]
\tikzstyle{connection}=
  [draw=lightgray,thin]
\tikzstyle{annotation}=
  [rectangle,rounded corners,draw=lightgray,fill=white,fill opacity=0.7,text opacity=1]

\tikzstyle{fmap}=
  [perceptron_input,rectangle,draw=blue!25!white]

\tikzstyle{box}=
  [rectangle,rounded corners,minimum width=3mm]

\tikzstyle{kernel}=
  [draw=gray,fill=blue!10!white]
\tikzstyle{bb}=
  [inner sep=0,outer sep=0]

\tikzstyle{unet_l1}=[]
\tikzstyle{unet_l2}=
  [scale=0.7,transform shape]
\tikzstyle{unet_l3}=
  [scale=0.5,transform shape]
\tikzstyle{unet_l4}=
  [scale=0.5,transform shape]

\tikzstyle{unet_annotation}=
  [circle,fill=white,draw=black,text=black,minimum width=4mm]

\tikzstyle{conv_pass}=
  [single arrow,shading=axis,left color=blue!50!white, right color=blue!10!white,draw=blue!50!white!70!black, inner sep=0]

\tikzstyle{copy_pass}=
  [single arrow,shading=axis,left color=purple!50!white, right color=purple!10!white,draw=purple!50!white!70!black, inner sep=0]

\tikzstyle{max_pool_pass}=
  [single arrow,shading=axis,left color=orange!50!white, right color=orange!10!white,draw=orange!50!white!70!black, inner sep=0]

\tikzstyle{upsampling_pass}=
  [single arrow,shading=axis,left color=brown!50!white, right color=brown!10!white,draw=brown!50!white!70!black, inner sep=0]

\tikzstyle{frame}=
  [rectangle,draw,fill=black]

\tikzstyle{pointer}=
  [->,thick,funkey_highlight]

\tikzstyle{gp_node}=
  [circle,draw,fill=gray!10!white,minimum width=0.5cm]
\tikzstyle{gp_annotation}=
  [rectangle,rounded corners,draw=gray!20!white,fill=gray!10!white,minimum width=0.5cm]

\tikzstyle{gp_edge}=
  [->,>=stealth',thick]
\tikzstyle{gp_request_edge}=
  [->,>=stealth',color=blue!50!white]
\tikzstyle{gp_batch_edge}=
  [->,>=stealth',color=orange!50!white]

\tikzstyle{gp_request}=
  [rectangle,rounded corners,fill=white,draw=gray]

\usepackage{pgfplotstable}
\usetikzlibrary{pgfplots.groupplots}
\usepgfplotslibrary{statistics}
\usepgfplotslibrary{patchplots}

\pgfplotsset{compat=1.16}
\pgfplotsset{
  emphasize/.code args={#1:#2with#3}{
    \pgfplotsextra{
      \draw[fill=#3] ({axis cs: 0,#1} -| {axis description cs:0,0}) 
       rectangle ({axis cs:0.4,#2} -| {axis description cs:1,0});
    }
  }
}

\pgfplotsset{
  errors/.style={
    stack plots=y,
    area style,
    enlarge x limits=false,
    xmajorgrids=true,
    ymajorgrids=true,
    yminorgrids=true,
    legend reversed
  }
}

\pgfplotsset{
  discard if not/.style 2 args={
    x filter/.code={
      \edef\tempa{\thisrow{#1}}
      \edef\tempb{#2}
      \ifx\tempa\tempb
      \else
        \def\pgfmathresult{inf}
      \fi
    }
  }
}

\pgfplotsset{
  discard if not both/.style args={#1 is #2 and #3 is #4}{
    x filter/.code={
      \edef\tempa{\thisrow{#1}}
      \edef\tempb{#2}
      \edef\tempc{\thisrow{#3}}
      \edef\tempd{#4}
      \ifx\tempa\tempb
        \ifx\tempc\tempd
        \else
          \def\pgfmathresult{inf}
        \fi
      \else
        \def\pgfmathresult{inf}
      \fi
    }
  }
}

\pgfplotsset{
  discard if not all three/.style args={#1 is #2 and #3 is #4 and #5 is #6}{
    x filter/.code={
      \edef\tempa{\thisrow{#1}}
      \edef\tempb{#2}
      \edef\tempc{\thisrow{#3}}
      \edef\tempd{#4}
      \edef\tempe{\thisrow{#5}}
      \edef\tempf{#6}
      \ifx\tempa\tempb
        \ifx\tempc\tempd
          \ifx\tempe\tempf
          \else
            \def\pgfmathresult{inf}
          \fi
        \else
          \def\pgfmathresult{inf}
        \fi
      \else
        \def\pgfmathresult{inf}
      \fi
    }
  }
}

\makeatletter
\def\parsenode[#1]#2\pgf@nil{%
    \tikzset{label node/.style={#1}}
    \def\nodetext{#2}
}

\tikzset{
    add node at x/.style 2 args={
        name path global=plot line,
        /pgfplots/execute at end plot visualization/.append={
                \begingroup
                \@ifnextchar[{\parsenode}{\parsenode[]}#2\pgf@nil
            \path [name path global = position line #1-1]
                ({axis cs:#1,0}|-{rel axis cs:0,0}) --
                ({axis cs:#1,0}|-{rel axis cs:0,1});
            \path [xshift=1pt, name path global = position line #1-2]
                ({axis cs:#1,0}|-{rel axis cs:0,0}) --
                ({axis cs:#1,0}|-{rel axis cs:0,1});
            \path [
                name intersections={
                    of={plot line and position line #1-1},
                    name=left intersection
                },
                name intersections={
                    of={plot line and position line #1-2},
                    name=right intersection
                },
                label node/.append style={pos=1}
            ] (left intersection-1) -- (right intersection-1)
            node [label node]{\nodetext};
            \endgroup
        }
    },
    add node at y/.style 2 args={
        name path global=plot line,
        /pgfplots/execute at end plot visualization/.append={
                \begingroup
                \@ifnextchar[{\parsenode}{\parsenode[]}#2\pgf@nil
            \path [name path global = position line #1-1]
                ({axis cs:0,#1}-|{rel axis cs:0,0}) --
                ({axis cs:0,#1}-|{rel axis cs:1,1});
            \path [yshift=1pt, name path global = position line #1-2]
                ({axis cs:0,#1}-|{rel axis cs:0,0}) --
                ({axis cs:0,#1}-|{rel axis cs:1,1});
            \path [
                name intersections={
                    of={plot line and position line #1-1},
                    name=left intersection
                },
                name intersections={
                    of={plot line and position line #1-2},
                    name=right intersection
                },
                label node/.append style={pos=1}
            ] (left intersection-1) -- (right intersection-1)
            node [label node] {\nodetext};
            \endgroup
        }
    }
}
\makeatother

\newcommand{\findmax}[3]{%
    \pgfplotstablevertcat{\datatable}{#1}%
    \pgfplotstablecreatecol[%
    create col/expr={%
    \pgfplotstablerow%
    }]{rownumber}\datatable%
    \pgfplotstablesave[skip rows between index={0}{1}]{\datatable}{tmp.table}%
    \pgfplotstableread{tmp.table}\datatableskipped%
    \pgfplotstablesort[sort key={#2},sort cmp={float >}]{\sorted}{\datatableskipped}%
    \pgfplotstablegetelem{0}{rownumber}\of{\sorted}%
    \pgfmathtruncatemacro#3{\pgfplotsretval}%
    \pgfplotstableclear{\datatable}%
    \pgfplotstableclear{\datatableskipped}%
}%
\newcommand{\findmaxx}[3]{%
    \pgfplotstablevertcat{\datatablee}{#1}%
    \pgfplotstablecreatecol[%
    create col/expr={%
    \pgfplotstablerow%
    }]{rownumber}\datatablee%
    \pgfplotstablesave[skip rows between index={0}{1}]{\datatablee}{tmp.table}%
    \pgfplotstableread{tmp.table}\datatableskippedd%
    \pgfplotstablesort[sort key={#2},sort cmp={float >}]{\sorted}{\datatableskippedd}%
    \pgfplotstablegetelem{0}{rownumber}\of{\sorted}%
    \pgfmathtruncatemacro#3{\pgfplotsretval}%
    \pgfplotstableclear{\datatablee}%
    \pgfplotstableclear{\datatableskippedd}%
}%
\newcommand{\findmaxxx}[3]{%
    \pgfplotstablevertcat{\datatableee}{#1}%
    \pgfplotstablecreatecol[%
    create col/expr={%
    \pgfplotstablerow%
    }]{rownumber}\datatableee%
    \pgfplotstablesave[skip rows between index={0}{1}]{\datatableee}{tmp.table}%
    \pgfplotstableread{tmp.table}\datatableskippeddd%
    \pgfplotstablesort[sort key={#2},sort cmp={float >}]{\sorted}{\datatableskippeddd}%
    \pgfplotstablegetelem{0}{rownumber}\of{\sorted}%
    \pgfmathtruncatemacro#3{\pgfplotsretval}%
    \pgfplotstableclear{\datatableee}%
    \pgfplotstableclear{\datatableskippeddd}%
}%
\newcommand{\findmaxxxx}[3]{%
    \pgfplotstablevertcat{\datatableeee}{#1}%
    \pgfplotstablecreatecol[%
    create col/expr={%
    \pgfplotstablerow%
    }]{rownumber}\datatableeee%
    \pgfplotstablesave[skip rows between index={0}{1}]{\datatableeee}{tmp.table}%
    \pgfplotstableread{tmp.table}\datatableskippedddd%
    \pgfplotstablesort[sort key={#2},sort cmp={float >}]{\sorted}{\datatableskippedddd}%
    \pgfplotstablegetelem{0}{rownumber}\of{\sorted}%
    \pgfmathtruncatemacro#3{\pgfplotsretval}%
    \pgfplotstableclear{\datatableeee}%
    \pgfplotstableclear{\datatableskippedddd}%
}%

\pgfplotstableset{%
    highlight row max/.code 2 args={%
        \pgfmathtruncatemacro\rowindex{#2-1}%
        \pgfplotstabletranspose{\transposed}{#1}%
        \findmax{\transposed}{\rowindex}{\maxval}%
        \edef\setstyles{\noexpand\pgfplotstableset{%
                every row \rowindex\space column \maxval\noexpand/.style={%
                    postproc cell content/.append style={%
                        /pgfplots/table/@cell content/.add={$\noexpand\bf}{$}%
                    },%
                }%
            }%
        }\setstyles%
    },%
    highlight row maxx/.code 2 args={%
        \pgfmathtruncatemacro\rowindex{#2-1}%
        \pgfplotstabletranspose{\transposed}{#1}%
        \findmaxx{\transposed}{\rowindex}{\maxval}%
        \edef\setstyles{\noexpand\pgfplotstableset{%
                every row \rowindex\space column \maxval\noexpand/.style={%
                    postproc cell content/.append style={%
                        /pgfplots/table/@cell content/.add={$\noexpand\bf}{$}%
                    },%
                }%
            }%
        }\setstyles%
    },%
    highlight row maxxx/.code 2 args={%
        \pgfmathtruncatemacro\rowindex{#2-1}%
        \pgfplotstabletranspose{\transposed}{#1}%
        \findmaxxx{\transposed}{\rowindex}{\maxval}%
        \edef\setstyles{\noexpand\pgfplotstableset{%
                every row \rowindex\space column \maxval\noexpand/.style={%
                    postproc cell content/.append style={%
                        /pgfplots/table/@cell content/.add={$\noexpand\bf}{$}%
                    },%
                }%
            }%
        }\setstyles%
    },%
    highlight row maxxxx/.code 2 args={%
        \pgfmathtruncatemacro\rowindex{#2-1}%
        \pgfplotstabletranspose{\transposed}{#1}%
        \findmaxxxx{\transposed}{\rowindex}{\maxval}%
        \edef\setstyles{\noexpand\pgfplotstableset{%
                every row \rowindex\space column \maxval\noexpand/.style={%
                    postproc cell content/.append style={%
                        /pgfplots/table/@cell content/.add={$\noexpand\bf}{$}%
                    },%
                }%
            }%
        }\setstyles%
    }%
}%

\makeatletter
\long\def\pgfplotstabletypeset@opt@collectarg[#1]#2{%

    \pgfplotstable@isloadedtable{#2}%
        {\pgfplotstabletypeset@opt@[#1]{#2}}%
        {\pgfplotstabletypesetfile@opt@[#1]{#2}}%
}
\makeatother

\newcommand{\getzoomfactor}{%
\pgfgettransformentries{\myxscale}{\@tempa}{\@tempa}{\myyscale}{\@tempa}{\@tempa}
\gdef\zoomfactor{\myxscale}
}

\makeatletter
\newlength{\hlheatmap@width}
\newlength{\hlheatmap@height}
\define@key{hlheatmap}{width}{\setlength\hlheatmap@width{#1}}
\define@key{hlheatmap}{height}{\setlength\hlheatmap@height{#1}}
\define@key{hlheatmap}{colorbar}{\def\hlheatmap@colorbar{#1}}
\define@key{hlheatmap}{ylabels}{\def\hlheatmap@ylabels{#1}}
\define@key{hlheatmap}{bgcolor}{\def\hlheatmap@bgcolor{#1}}
\define@key{hlheatmap}{perneuronlabels}{\def\hlheatmap@perneuronlabels{#1}}
\define@key{hlheatmap}{hldata}{\def\hlheatmap@hldata{#1}}
\define@key{hlheatmap}{noexport}{\def\hlheatmap@noexport{#1}}
\def\hlheatmap@true{true}
\def\hlheatmap@false{false}

\makeatother

\let\todonote\todo
\renewcommand{\todo}[1]{\todonote[color=orange!50!white]{\footnotesize #1}\xspace}

\usepackage{grfext}
\PrependGraphicsExtensions*{.jpg,.JPG}

\makeatletter
\newcolumntype{d}{D{.}{.}{-1} }
\newcolumntype{B}[3]{>{\boldmath\DC@{#1}{#2}{#3} }c<{\DC@end} }
\makeatother

\def\mathlet#1#2{\pgfmathparse{#2}\let#1\pgfmathresult}

\def\scale#1#2{\tikz{\node[scale=#1,transform shape,inner sep=0]{#2};}}

\def\figref#1{Fig.~\ref{#1}}

\def\secref#1{Section~\ref{#1}}
\def\tabref#1{Table~\ref{#1}}
\def\eqref#1{(\ref{#1})}

\newcommand{\image}[2][]{\tikz{\node[inset]{\includegraphics[#1]{#2}};}}
\newlength{\volumethickness}
\newcommand{\drawcube}[6][]{
  \setlength{\volumethickness}{#6}

  \mathlet\b{int(#3)} %
  \mathlet\e{int((#4-1))} %

  \mathlet\reps{int(10)} %
  \mathlet\t{int(#5*\reps)} %

  \foreach \idx in {\b,...,\e}{
    \foreach \r in {0,...,\reps}{
      \mathlet\i{int(\idx*\reps+\r)}
      \mathlet\off{-\i/\t}
      \node[inner sep=0,outer sep=0] (img_\i) at (\off\volumethickness, \off\volumethickness) {
        \includegraphics[#1]{#2\idx}
      };
    }
  }

  \mathlet\rb{int(\b*\reps)}
  \mathlet\re{int((\e+1)*\reps)}
  \path[fill opacity=0.4,fill=white,draw=black,very thin,draw opacity=0.2]
    (img_\rb.north west) --
    (img_\re.north west) --
    (img_\re.north east) --
    (img_\rb.north east) --
    cycle;
  \path[fill opacity=0.4,fill=black,draw=black,very thin,draw opacity=0.2]
    (img_\rb.north east) --
    (img_\re.north east) --
    (img_\re.south east) --
    (img_\rb.south east) --
    cycle;
}

\makeatletter
\define@key{volume}{width}{\def\volume@width{#1}}
\define@key{volume}{thickness}{\def\volume@thickness{#1}}
\define@key{volume}{center offset}{\def\volume@centeroffset{#1}}
\newcommand{\volume}[3][]{
  \setkeys{volume}{width=2cm,thickness=5mm,center offset=1.25cm,#1}
  \tikz{
    \mathlet\m{int(#3/2)} %
    \mathlet\mn{int(\m+1)} %
    \drawcube[width=\volume@width]{#2}{0}{\m}{#3}{\volume@thickness}
    \begin{scope}[xshift=-\volume@centeroffset,yshift=\volume@centeroffset]
      \drawcube[width=\volume@width]{#2}{\m}{\mn}{#3}{\volume@thickness}
    \end{scope}
    \drawcube[width=\volume@width]{#2}{\mn}{#3}{#3}{\volume@thickness}
  }
}
\makeatother

\begin{document}

\title{Discriminative Attribution from Counterfactuals}

\author{%
  Nils Eckstein\\
  HHMI Janelia Research Campus\\
  Ashburn, VA 20147\\
  \& ETH Zürich\\
  Zürich, Switzerland\\
  \texttt{nils.eckstein@googlemail.com}\\
  \And
  Alexander S. Bates\\
  Neurobiology Division\\
  MRC Laboratory of Molecular Biology\\
  Cambridge, United Kingdom\\
  \And
  Gregory S.X.E. Jefferis\\
  Neurobiology Division\\
  MRC Laboratory of Molecular Biology\\
  Cambridge, United Kingdom\\
  \And
  Jan Funke\\
  HHMI Janelia Research Campus\\
  Ashburn, VA 20147\\
  \texttt{funkej@janelia.hhmi.org}
}

\date{}
\maketitle
\thispagestyle{empty}

\begin{abstract}
We present a method for neural network interpretability by combining feature
attribution with counterfactual explanations to generate attribution maps that
highlight the most discriminative features between pairs of classes.
We show that this method can be used to quantitatively evaluate the performance
of feature attribution methods in an objective manner, thus preventing
potential observer bias.
We evaluate the proposed method on three diverse datasets, including a
challenging artificial dataset and real-world biological data.
We show quantitatively and qualitatively that the highlighted features are
substantially more discriminative than those extracted using conventional
attribution methods and argue that this type of explanation is better suited
for understanding fine grained class differences as learned by a deep neural
network.
\end{abstract}

\section{Introduction}

Machine Learning---and in particular Deep Learning---continues to see increased
adoption in crucial aspects of society such as industry, science, and
healthcare.
  As such, it impacts human lives in significant ways. Consequently, there is a
  need for understanding how these systems work and how they make predictions
  in order to increase user trust, limit the perpetuation of societal biases,
  ensure correct function, or even to gain scientific knowledge. However, due
  to the large numbers of parameters and non-linear interactions between input
  and output, deep neural networks (DNNs) are generally hard to interpret. In
  particular, it is not clear which input features influence the output of a
  DNN.

A popular approach for explaining DNN predictions is provided by so-called
feature attribution methods.
  These methods output the importance of each input feature w.r.t.\ the output
  of the DNN. In the case of image classification---the primary focus of this
  work---the output is a heatmap over input pixels, highlighting and ranking
  areas of importance. A large number of approaches for feature attribution
  have been proposed in recent years (for a recent review
  see~\cite{samek2021explaining} and related work below). Although those have
  been used successfully to interpret model behavior for some applications,
  there is still debate about the effectiveness, accuracy, and trustworthiness
  of these
  approaches~\citep{kindermans2019reliability,adebayo2018sanity,ghorbani2019interpretation,alvarez2018robustness}.
  In addition, objectively evaluating feature attribution methods remains a
  difficult task~\citep{samek2016evaluating,hooker2018benchmark}.

A complementary approach for explaining DNN decisions are so called
counterfactual
explanations~\citep{martens2014explaining,wachter2017counterfactual}.
  In contrast to feature importance estimation, counterfactual approaches
  attempt to explain a DNN output by presenting the user with another input
  that is close to the original input, but changes the classification decision
  of the DNN to another class. For humans, this representation is natural and
  it provides an intuitive means for elucidating DNN behaviour. 

While counterfactual explainability methods have seen increased adoption in
structured data domains, they are comparatively less popular for image data,
where feature attribution methods arguably remain the dominant tool for
practitioners.
  The popularity of feature importance methods is partly driven
  by their ease of use, availability in popular Deep Learning frameworks, and
  intuitive outputs in the form of pixel heatmaps. In contrast, generating
  counterfactual explanations typically involves an optimization procedure that
  needs to be carefully tuned in order to obtain a counterfactual with the
  desired properties. This process can be computationally expensive and does,
  in general, not allow for easy computation of attribution
  maps~\citep{verma2020counterfactual}.

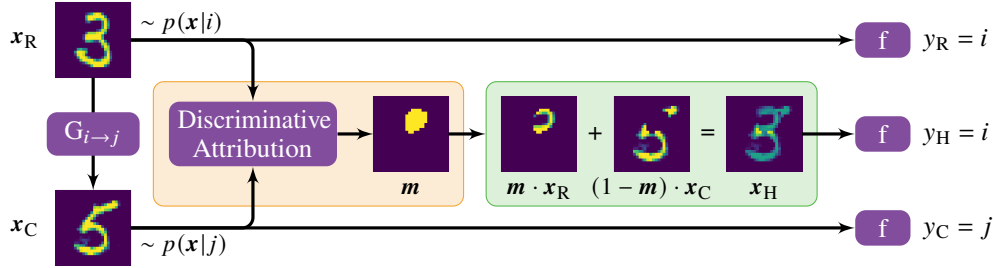
\begin{figure}
  \centerline{\begin{tikzpicture}

  \node[image]            (x_a)          {\image[width=10mm]{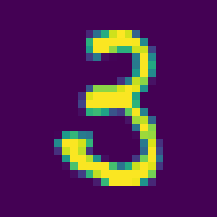}};
  \node[image,below=20mm] (x_b) at (x_a) {\image[width=10mm]{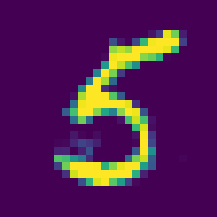}};
  \node[left=1mm] at (x_a.west) {\footnotesize$\xo$};
  \node[left=1mm] at (x_b.west) {\footnotesize$\xc$};
  \node[right,anchor=north west] at (x_a.north east) {\footnotesize$\sim p(\x|i)$};
  \node[right,anchor=south west] at (x_b.south east) {\footnotesize$\sim p(\x|j)$};
  \draw[arrow]
    (x_a)
    --
    node[pos=0.5,method,name=g_b,text width=10mm] {$\text{G}_{i\rightarrow j}$}
    (x_b);

  \node[method,text width=20mm,right=10mm] (dac) at (g_b) {Discriminative Attribution};
  \draw[arrow,rounded corners] (x_a) -- ($(x_a-|dac)$) -- (dac);
  \draw[arrow,rounded corners] (x_b) -- ($(x_b-|dac)$) -- (dac);
  \node[image,right=16mm] (m) at (dac) {\image[width=10mm]{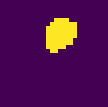}};
  \node[below=8mm,anchor=base,inner sep=0] (ann_m) at (m) {\footnotesize$\mask$};
  \draw[arrow] (dac) -- (m);

  \node[image,right=12mm] (m_x_a) at (m) {\image[width=10mm]{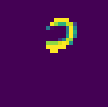}};
  \node[below=8mm,anchor=base] at (m_x_a) {\footnotesize$\mask\cdot\xo$};

  \node[image,right=10mm] (m_x_b) at (m_x_a) {\image[width=10mm]{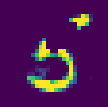}};
  \node[below=8mm,anchor=base] at (m_x_b) {\footnotesize$(1-\mask)\cdot\xc$};

  \node at ($(m_x_a)!.5!(m_x_b)$) {+};

  \node[image,right=10mm] (x_m) at (m_x_b) {\image[width=10mm]{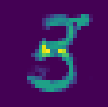}};
  \node[below=8mm,anchor=base,inner sep=0] (ann_x_m) at (x_m) {\footnotesize$\xh$};

  \node at ($(m_x_b)!.5!(x_m)$) {=};

  \begin{pgfonlayer}{background}
    \node[
      rounded corners,
      draw=funkey_color_2,
      fill=funkey_color_2,
      fill opacity=0.2,
      inner sep=2mm,
      fit=(dac)(m)($(ann_x_m-|m)$)] (dac_m) {};
    \node[
      rounded corners,
      draw=funkey_color_3,
      fill=funkey_color_3,
      fill opacity=0.2,
      inner sep=2mm,
      fit=(m_x_a)(x_m)($(ann_x_m-|x_m)$)] (mix) {};
  \end{pgfonlayer}
  \draw[arrow] (m) -- ($(mix.west|-m)$);

  \node[method,right=12mm] (c_m) at (x_m) {f};
  \node[method] (c_a) at ($(c_m|-x_a)$) {f};
  \node[method] (c_b) at ($(c_m|-x_b)$) {f};
  \draw[arrow] (x_a) -- (c_a);
  \draw[arrow] (x_b) -- (c_b);
  \draw[arrow] (x_m) -- (c_m);

  \node[right=4mm] (y_a) at (c_a) {$\yo = i$};
  \node[right=4mm] (y_m) at (c_m) {$\yh = i$};
  \node[right=4mm] (y_b) at (c_b) {$\yc = j$};

\end{tikzpicture}}
  \caption{
    Overview of the proposed method:
    An input image $\xo$ of class $i$ is converted through an independently trained cycle-GAN
    generator $G_{i\rightarrow j}$ into a counterfactual image $\xc$ of class
    $j$, such that the classifier $f$ we wish to interpret predicts $\yo=i$ and
    $\yc=j$.
    The Discriminative Attribution method then searches for the minimal
    mask $\mask$, such that copying the most discriminative parts of the
    real image $\xo$ into the counterfactual $\xc$ (resulting in the hybrid
    $\xh$) is again classified as $\yh=i$.
  }
  \label{fig:overview}
\end{figure}

To address these issues, we present a simple method that bridges the gap
between counterfactual explainability and feature importance for image
classification by building attribution maps from counterfactuals (DAC:
Discriminative Attribution from Counterfactuals, see~\figref{fig:overview} for
a visual summary).
  Crucially, our method can be used to quantitatively evaluate the attribution
  in an objective manner on a target task, a missing feature in current
  attribution methods.
  We use a cycle-GAN~\citep{zhu2017unpaired} to translate real images $\xo$
  of class $i$ to counterfactual images $\xc$ of class $j\neq i$, where we
  validate that the translation has been successful by confirming that
  $f(\xo)=i$ and $f(\xc)=j$, where $f$ is the classifier to interpret.
  We repurpose a set of common attribution methods by introducing their
  discriminative counterparts, which are then able to derive attribution maps
  from the paired real and counterfactual image.
  We show that this approach is able to generate sparse, high quality feature
  attribution maps that highlight the most discriminative features in the
  real and counterfactual image more precisely than standard attribution
  methods.
  Furthermore, subsequent thresholding of the attribution map allows us to
  extract binary masks of the features and quantify their discriminatory power
  by performing an intervention and replacing the highlighted pixels in the
  counterfactual with the corresponding pixels in the real image. The
  difference in output classification score of this hybrid image, compared to
  the real image classification, then quantifies the importance of the swapped
  features.
  We validate our method on a set of three diverse tasks, including a
  challenging artificial dataset, a real world biological dataset (where a DNN
  solves a task human experts can not), and MNIST. For all three datasets we show
  quantitatively and qualitatively that our method outperforms all considered
  attribution methods in identifying key discriminatory features between the
  classes.
  Source code and datasets are publicly available at
  \url{https://dac-method.github.io}.

\section{Related Work}

Recent years have seen a large number of contributions addressing the problem
of interpretability of DNNs.
  These can be broadly distinguished by the type of explanation they provide,
  either \emph{local} or \emph{global}.
  Methods for local interpretability provide an explanation for every input,
  highlighting the reasons why a particular input is assigned to a certain
  class by the DNN.
  Global methods attempt to distill the DNN in a representation that is easier
  to understand for humans, such as decision trees.
  One can further distinguish between interpretability methods that are
  post-hoc, i.e., applicable to every DNN after it has been trained, and those
  methods that require modifications to existing architectures to perform
  interpretable classification as part of the model.
  In this work we focus on a specific class of local, post-hoc approaches to
  DNN interpretability for image classification, so-called feature importance
  estimation methods.

\paragraph{Attribution Methods for Image Classification}
Even in this restricted class of approaches there is a large variety of
methods~\citep{ribeiro2016model,lundberg2017unified,baehrens2010explain,bach2015pixel,zintgraf2017visualizing,selvaraju2017grad,Sundararajan2017,simonyan2014deep,zeiler2014visualizing,kindermans2017learning,montavon2017explaining,fong2017interpretable,dabkowski2017real,zhang2016understanding,shrikumar2017learning,shrikumar2016not,smilkov2017smoothgrad}\todo{Find more current}.
  They have in common that they aim to highlight the most important features
  that contributed to the output classification score for a particular class,
  generating a heatmap indicating the influence of input pixels and features on
  the output classification. Among those, of particular interest to the work
  presented here are \textit{baseline} feature importance methods, which
  perform feature importance estimation with reference to a second input. Those
  methods gained popularity, as they assert \textit{sensitivity} and
  \textit{implementation invariance}~\citep{Sundararajan2017,shrikumar2016not}.
  The baseline is usually chosen to be the zero image as it is assumed to
  represent a neutral input.

\paragraph{Counterfactual Interpretability}
Another body of literature that is relevant to the presented work are
counterfactual interpretability methods first proposed
by~\cite{martens2014explaining}.
  Since then, the standard approach for generating counterfactuals broadly
  follows the procedure proposed by~\cite{wachter2017counterfactual}, in which
  the counterfactual is found as a result of an optimization aiming to maximize
  output differences while minimizing input differences between the real image
  $\xo$ and the counterfactual $\xc$:
  \begin{equation}
      \xc =
        \underset{\x}{\text{argmin}}
        \;\;
        L_i(\xo,\x) - L_o(f(\xo), f(\x))
      \label{eq:cf}
  \end{equation}
  with $L_i$ and $L_o$ some loss that measures the distance between inputs and
  outputs, respectively, and $f$ the classifier in question.
  However, optimizing this objective can be problematic because it contains
  competing losses and does not guarantee that the generated counterfactual
  $\xc$ is part of the data distribution $p(\x)$. Current approaches try to
  remedy this by incorporating additional regularizers in the
  objective~\citep{liu2019generative,verma2020counterfactual}, such as
  adversarial losses that aim to ensure that the counterfactual $\xc$ is not
  distinguishable from a sample $\x \sim
  p(\x)$~\citep{barredo2020plausible,liu2019generative}.
  However, this does not address the core problem of competing objectives and
  will result in a compromise between obtaining in-distribution samples,
  maximizing class differences, and minimizing input differences. Interpreting
  the presented work in this context, we circumvent this issue by dropping the
  input similarity loss in the generation of counterfactuals and instead
  enforce similarity post-hoc, similar to the strategy used
  by~\cite{mothilal2020explaining}.

A closely related work addressing counterfactual interpetability is the method
presented by~\cite{narayanaswamy2020scientific}.
  Similar to ours, this method uses a cycle-GAN to generate counterfactuals for
  DNN interpretability. However, this method differs in that the cycle-GAN is
  applied multiple times to a particular input in order to increase the visual
  differences in the real and counterfactual images for hypothesis generation.
  Subsequently, the found features are confirmed by contrasting the original
  classifiers performance with one that is trained on the discovered features.
  Similar to other previous methods, this does not lead to attribution maps or
  an objective evaluation of feature importance.

\paragraph{Attribution and Counterfactuals}
Closest to our approach is the work by~\cite{wang2020scout}, which proposes to
combine attribution and counterfactual explanations.
  This work introduces a novel family of so-called \textit{discriminative
  explanations} that also leverage attribution on a real and counterfactual
  image in addition to confidence scores from the classifier to derive
  attributions for the real and counterfactual image that show highly
  discriminative features. In contrast to our work, this approach requires
  calculation of three different attribution maps, which are subsequently
  combined to produce a discriminative explanation. In addition, this method
  does not generate new counterfactuals using a generative model, but instead
  selects a real image from a different class. On one hand this is advantageous
  because it does not depend on the generator's performance, but on the other
  hand this does not allow creating hybrid images for the evaluation of
  attribution maps.

Another relevant work is presented by~\cite{goyal2019counterfactual}.
  Similar to our work, the authors devise a method to generate counterfactual
  visual explanations by searching for a feature pair in two real images of
  different classes that, if swapped, influences the classification decision.
  To this end, they propose an optimization procedure that searches for the
  best features to swap, utilizing the networks feature representations. In
  contrast to our work, the usage of real (instead of generated)
  counterfactuals can lead to more artifacts during the replacement of features. In addtion, our work
  supports the generation of attribution maps and features a procedure for the
  quantitative evaluation of the explanations.

\paragraph{Attribution Evaluation}
Our work differs notably from the current state of the art as it enables
quantitative evaluation of the generated attributions by copy-pasting features
from a paired image set (the real and the counterfactual).
  Prior work evaluated the importance of highlighted features by removing
  them~\citep{samek2016evaluating}. However, it has been noted that this
  strategy is problematic because it is unclear whether any observed
  performance degradation is due to the removal of relevant features or because
  the new sample comes from a different distribution. As a result, strategies
  to remedy this issue have been proposed, for example by retraining
  classifiers on the modified samples~\citep{hooker2018benchmark}. Instead of
  removing entire features, in this work we replace them with their
  corresponding counterfactual features.

\section{Method}

The method we propose combines counterfactual interpretability with
discriminative attribution methods to find and highlight the most important
features between images of two distinct classes $i$ and $j$, given a pretrained
classifier $f$.
  For that, we first generate for a given input image $\xo$ of class $i$ a
  counterfactual image $\xc$ of class $j$. We then use a \emph{discriminative}
  attribution method to find the attribution map of the classifier for this
  pair of images. As we will show qualitatively and quantitatively in
  \secref{sec:results}, using paired images results in attribution maps of
  higher quality. Furthermore, the use of a counterfactual image gives rise to
  an objective evaluation procedure for attribution maps.

In the next sections we describe (1) our choice for generating counterfactual
images, (2) the derivation of discriminative attribution methods from existing
baseline attribution methods, and (3) how to use counterfactual images to
evaluate attribution maps.
  We denote with $f$ a pretrained classifier with $N$ output classes, input
  images $\x\in\mathbb{R}^{h \times w}$, and output vector $f(\x)=\vct{y} \in
  [0,1]^N$ with $\sum_i y_i=1$.

\subsection{Creation of Counterfactuals}

We train a cycle-GAN~\citep{zhu2017unpaired} for each pair of image classes
$i\neq j\in\{0,1,...,N-1\}$, which enables translation of images of class $i$
into images of class $j$ and vice versa.
  We perform this translation for each image of class $i$ and each target class
  $j\neq i$ to obtain datasets of paired images $D_{i\rightarrow j} =
  \left\{(\xo^k, \xc^k)|k=1,\ldots,n(i)\right\}$, where $\xo^k$ denotes the
  $k$th real image of class $i$ and $\xc^k$ its counterfactual of class $j$.
  We then test for each image in the dataset whether the translation was
  successful by classifying the counterfactual image $\xc$ and reject a sample
  pair whenever $f(\xc)_j<\theta$, with $\theta$ a threshold parameter (in the
  rest of this work we set $\theta=0.8$, except otherwise specified). \todo{are
  there different values for $\theta$?}

This procedure results in a dataset of paired images, where the majority of the
differences between an image pair is expected to be relevant for the
classifiers decision, i.e., we retain formerly present non-discriminatory
distractors such as orientation, lighting, or background.
  We encourage that the translation makes as little changes as necessary by
  choosing a Res-Net~\citep{he2016deep} architecture for the cycle-GAN
  generator, which is able to trivially learn the identity function.

\subsection{Discriminative Attribution from Counterfactuals}

The datasets $D_{i\rightarrow j}$ are already useful to visualize
data-intrinsic class differences (see \figref{fig:res_qual} for
examples).
  However, we wish to understand which input features the classifier $f$ makes
  use of.
  Specifically, we are interested in finding the smallest binary mask $\mask$,
  such that swapping the contents of $\xc$ with $\xo$ within this mask changes
  the classification under $f$.

To find $\mask$, we repurpose existing attribution methods that are amendable
to be used with a reference image.
  The goal of those methods is to produce attribution maps $\am$, which we
  convert into a binary mask via thresholding.
  A natural choice for our purposes are so-called \emph{baseline attribution
  methods}, which derive attribution maps by contrasting an input image with a
  baseline sample (e.g., a zero image).
  In the following, we review suitable attribution methods and derive
  discriminative versions that use the counterfactual image as their baseline.
  We will denote the discriminative versions with the prefix $\pred$.

\subsubsection{Input * Gradients}

One of the first and simplest attribution methods is \textit{Input * Gradients}
(INGRADS)~\citep{shrikumar2016not,simonyan2014deep}, which is motivated by the
first order Taylor expansion of the output class with respect to the input
around the zero point:
  \begin{equation}
    \text{INGRADS}(\x) = |\nabla_{\x} f(\x)_i \cdot \x|
      \text{,}
  \end{equation}
  where $i$ is the class for which an attribution map is to be generated.
  We derive an explicit baseline version for the discriminatory attribution of
  the real $\xo$ and its counterfactual $\xc$ by choosing $\xc$ as the Taylor
  expansion point:
  \begin{equation}
    \pred\text{-INGRADS}(\xo, \xc) = |\nabla_{\x} f(\x)_j\Bigr|_{\x=\xc} \cdot (\xc - \xo)|
    \text{,}
  \end{equation}
  where $j$ is the classes of the counterfactual image.

\subsubsection{Integrated Gradients}

\textit{Integrated Gradients} (IG) is an explicit baseline attribution method,
where gradients are accumulated along the straight path from a baseline input
$\x_0$ to the input image $\x$ to generate the attribution
map~\citep{Sundararajan2017}.
  Integrated gradients along the $k$th dimension are given by:
  \begin{equation}
    \text{IG}_k(\x) =
      (\x - \x_0)_k \cdot
      \int_{\alpha = 0}^{1}
        \frac{\partial f (\x_0 + \alpha (\x - \x_0))_i}{\partial \x_k}
        d\alpha
    \text{.}
  \end{equation}
  We derive a discriminatory version of IG by replacing the baseline as
  follows:
  \begin{equation}
    \pred\text{-IG}_k(\xo, \xc) =
      (\xc - \xo)_k \cdot
      \int_{\alpha = 0}^{1}
      \frac{\partial f (\xo + \alpha (\xc - \xo))_j}{\partial {\x}_k}
        d\alpha
    \text{.}
  \end{equation}

\subsubsection{Deep Lift}

\textit{Deep Lift} (DL) is also an explicit baseline attribution method which
aims to compare individual neurons activations of an input w.r.t.\ a reference
baseline input~\citep{shrikumar2016not}.
  It can be expressed in terms of the gradient in a similar functional form to
  IG:
  \begin{equation}
    \text{DL}(\x) = (\x - \x_0) \cdot F_{DL}
    \text{,}
  \end{equation}
  where $F_{DL}$ is some function of the gradient of the output
  (see~\cite{ancona2018towards} for the full expression). The discriminative
  attribution we consider is simply:
  \begin{equation}
    \pred\text{-DL}(\xo, \xc) = (\xc - \xo) \cdot F_{DL}
    \text{.}
  \end{equation}

\subsubsection{GradCAM}

\textit{GradCAM} (GC) is an attribution method that considers the gradient
weighted activations of a particular layer, usually the last convolutional
layer, and propagates this value back to the input
image~\citep{selvaraju2017grad}.
We denote the activation of a pixel $(u, v)$ in layer $l$ with size $(h, w)$
and channel $k$ by $C^{l}_{k,u,v}$ and write the gradient w.r.t.\ the output
$\vct{y}$ as:
  \begin{equation}
    \nabla_{C^{l}_{k}}\vct{y} =
    (
      \frac{d\vct{y}}{dC^{l}_{k,0,0}},
      \frac{d\vct{y}}{dC^{l}_{k,1,0}},
      \frac{d\vct{y}}{dC^{l}_{k,2,0}},
      ...,
      \frac{d\vct{y}}{dC^{l}_{k,h,w}}
    )
  \end{equation}
  The original GC is then defined as:
  \begin{equation}
    \text{GC}(\x) =
      \text{ReLU}
      \left(
        \sum_k \nabla_{C_{k}}\vct{y} \cdot \vec{C}_{k}
      \right) =
      \text{ReLU}
      \left(
        \sum_{k}\sum_{u,v} \frac{d\vct{y}}{dC_{k,u,v}} C_{k,u,v}
      \right)
      =
      \text{ReLU}
      \left(
        \sum_k \alpha_k C_{k}
      \right)
    \text{,}
  \end{equation}
  where we ommitted the layer index $l$ for brevity.
  Each term $\frac{d\vct{y}}{dC_{k,u,v}} C_{k,u,v}$ is the contribution of
  pixel $u, v$ in channel $k$ to the output classification score $\vct{y}$
  under a linear model. GC utilizes this fact and projects the layer
  attribution from layer $l$ back to the input image, generating the final
  attribution map.

In contrast to the setting considered by GC, we have access to a matching pair
of real and counterfactual images $\xo$ and $\xc$.
  We extend GC to consider both feature maps $C^{\xo}_k$ and $C^{\xc}_k$ by
  treating GC as an implicit zero baseline method similar to INGRADS:
  \begin{equation}
    \pred\text{-GC}_{k}(\xo, \xc) =
      \frac{d\vct{y}_j}{dC_{k}} \Bigr|_{C=C^{\xc}_k}
      \left(
        C_{k}(\xc) - C_{k}(\xo)
      \right)
    \text{.}
  \end{equation}
  Averaging those gradients over feature maps $k$, and
  projecting the activations back to image space then highlights pixels that
  are most discriminative for a particular pair:
  \begin{equation}
    \pred\text{-GC}_{\text{P}}(\xo, \xc) =
      \left|\mathbb{P}\sum_{k} \pred\text{-GC}_k(\xo,\xc)\right|
    \text{,}
  \end{equation}
  where $\mathbb{P}$ is the projection matrix from feature space $C$ to input
  space $X$. Note that in contrast to GC, we use the absolute value of the
  output attribution, as we do not apply ReLU activations to layer
  attributions.
  \todo{I don't understand that}

Because feature maps can be of lower resolution than the input space, GC tends
to produce coarse attribution maps~\citep{selvaraju2017grad}.
  To address this issue it is often combined with \textit{Guided
  Backpropagation} (GBP), a method that uses the gradients of the output class
  w.r.t.\ the input image as the attribution
  map~\citep{springenberg2014striving}. During the backwards pass, all values
  $<0$ at each ReLU non-linearity are then discarded to only retain positive
  attributions.
  \todo{how is that related to the prev sentence?}

\textit{Guided GradCAM} (GGC) uses this strategy to sharpen the attribution of
GC via element-wise multiplication of the attribution
maps~\citep{selvaraju2017grad}.
  For the baseline versions we thus consider multiplication of
  $\pred\text{-GC}$ with the GBP attribution maps:
  \begin{align}
    \text{GBP}(\x) &= \nabla_{\x} f(\x)_i \;\;\text{, with} \nabla\text{ReLU} > 0 \\
    \text{GGC}(\x) &= \text{GC}(\x) \cdot \text{GBP}(\x) \\
    \pred\text{-GGC}(\xo, \xc) &= \pred\text{-GC}(\xo, \xc) \cdot \text{GBP}(\xo)
    \text{.}
  \end{align}

\subsection{Evaluation of Attribution Maps}
\label{sec:method:evaluation}

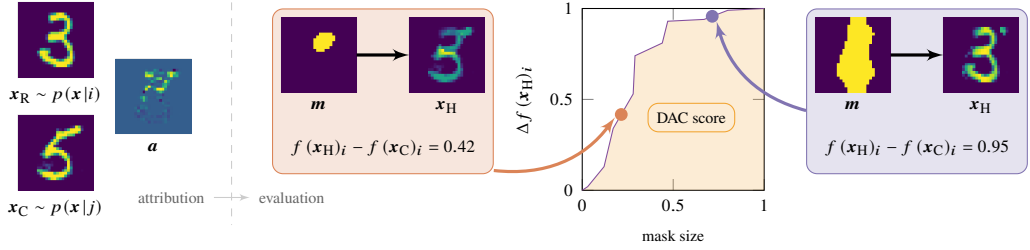
\begin{figure}
  \centerline{\begin{tikzpicture}
  \begin{axis}[
      xlabel={\tiny mask size},
      ylabel={\tiny$\Delta f(\xh)_i$},
      legend style={at={(0.98,0.02)},anchor=south east},
      xmin=0,
      ymin=0,
      xmax=1,
      ymax=1,
      xtick={0,0.5,1},
      ytick={0,0.5,1},
      x tick label style={font=\tiny},
      y tick label style={font=\tiny},
      width=4cm,
      height=4cm
    ]

    \addplot[
        draw=none,
        fill=funkey_color_2,
        fill opacity=0.2,
        mark=none]
      table [
        col sep=comma,
        x=mask_size,
        y=delta
      ] {figures/results_figures/data/dummy_dac_score.csv} \closedcycle;
    \addplot[
        draw=funkey_color_1,
        mark=none,
        add node at x={0.2}{
          [
            name=sample1,
            circle,
            inner sep=0.6mm,
            fill=funkey_color_4
          ]{}
        },
        add node at x={0.7}{
          [
            name=sample2,
            circle,
            inner sep=0.6mm,
            fill=funkey_color_5
          ]{}
        },
      ]
      table [
        col sep=comma,
        x=mask_size,
        y=delta
      ] {figures/results_figures/data/dummy_dac_score.csv};

    \coordinate (dac_score) at (axis cs: 0.6, 0.4);

  \end{axis}

  \node[image] (x_m_1) at (-3.5, 1.8) {\image[width=10mm]{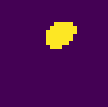}};
  \node[image,right=12mm] (x_h_1) at (x_m_1) {\image[width=10mm]{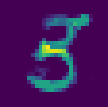}};
  \draw[arrow] (x_m_1) -- (x_h_1);
  \node[below] (ann_x_m_1) at (x_m_1.south) {\tiny$\mask$};
  \node[below] (ann_x_h_1) at (x_h_1.south) {\tiny$\xh$};
  \node[below=10mm] (c_1) at ($(x_m_1)!.5!(x_h_1)$) {\tiny$f(\xh)_i - f(\xc)_i = 0.42$};

  \begin{pgfonlayer}{background}
    \node[
      rounded corners,
      draw=funkey_color_4,
      fill=funkey_color_4,
      fill opacity=0.2,
      fit=(x_m_1)(x_h_1)(c_1)(ann_x_h_1)
    ] (box_sample1) {};
  \end{pgfonlayer}
  \draw[funkey_color_4,arrow] (box_sample1) to[bend right] (sample1);

  \node[image] (x_m_2) at (3.6, 1.8) {\image[width=10mm]{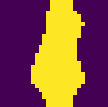}};
  \node[image,right=12mm] (x_h_2) at (x_m_2) {\image[width=10mm]{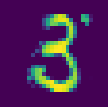}};
  \draw[arrow] (x_m_2) -- (x_h_2);
  \node[below] (ann_x_m_2) at (x_m_2.south) {\tiny$\mask$};
  \node[below] (ann_x_h_2) at (x_h_2.south) {\tiny$\xh$};
  \node[below=10mm] (c_2) at ($(x_m_2)!.5!(x_h_2)$) {\tiny$f(\xh)_i - f(\xc)_i = 0.95$};

  \begin{pgfonlayer}{background}
    \node[
      rounded corners,
      draw=funkey_color_5,
      fill=funkey_color_5,
      fill opacity=0.2,
      fit=(x_m_2)(x_h_2)(c_2)(ann_x_h_2)
    ] (box_sample2) {};
  \end{pgfonlayer}
  \draw[funkey_color_5,arrow] (box_sample2) to[bend left] (sample2);

  \node[image,left=30mm,yshift=2mm] (x_o) at (x_m_1) {\image[width=10mm]{figures/data/samples/figure_1/4_dl_diff_1_real}};
  \node[image,below=10mm] (x_c) at (x_o) {\image[width=10mm]{figures/data/samples/figure_1/4_dl_diff_1_fake}};
  \node[image,right=8mm] (a) at ($(x_c)!.5!(x_o)$) {\image[width=10mm]{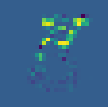}};
  \node[below] at (x_o.south) {\tiny$\xo\sim p(\x|i)$};
  \node[below] (ann_x_c) at (x_c.south) {\tiny$\xc\sim p(\x|j)$};
  \node[below] at (a.south) {\tiny$\am$};

  \coordinate (sep) at ($(a.east)!.5!(box_sample1.west)$);
  \draw[thin,dashed=1mm,funkey_lightgrey]
    ($(x_o.north-|sep)$)
    --
    node[left=8mm,pos=0.9,funkey_grey,name=att,anchor=base] {\tiny attribution}
    node[right=8mm,pos=0.9,funkey_grey,name=eval,anchor=base] {\tiny evaluation}
    ($(ann_x_c.south-|sep)$);
  \draw[arrow,thin,funkey_lightgrey] (att) -- (eval);

  \node[
      rounded corners,
      draw=funkey_color_2
    ] (dac) at (dac_score) {\tiny DAC score};

\end{tikzpicture}}
  \caption{Evaluation procedure for discriminative attribution methods:
    Given the real image $\xo$ of class $i$ and its counterfactual $\xc$ of
    class $j$, we generate a sequence of binary masks $\mask$ by applying
    different thresholds to the attribution map $\am$. Those masks are then
    used to generate a sequence of hybrid images $\xh$.
    The plot shows the change in classifier prediction $\Delta f(\xh)_i =
    f(\xh)_i - f(\xc)_i$ over the size of the mask $\mask$ (normalized between
    0 and 1).
    The DAC score is the area under the curve, i.e., a value between 0 and 1.
    Higher DAC scores are better and indicate that a discriminative attribution
    method found small regions that lead to the starkest change in
    classification.
  }
  \label{fig:method:evaluation}
\end{figure}

The discriminative attribution map $\am$ obtained for pair of images $(\xo,
\xc)$ can be used to quantify the causal effect of the attribution.
  Specifically, we can copy the area highlighted by $\am$ from the real image
  $\xo$ of class $i$ to the counterfactual image $\xc$ of class $j$, resulting
  in a hybrid image $\xh$.
  If the attribution accurately captures class-relevant features, we would
  expect that the classifier $f$ assigns a high probability to $\xh$ being of
  class $i$.

The ability to create those hybrid images is akin to an intervention, and has
two important practical implications:
  First, it allows us to find a minimal binary mask that captures the most
  class-relevant areas for a given input image.
  Second, we can compare the change in classification score for hybrids derived
  from different attribution maps. This allows us to compare different methods
  in an objective manner, following the intuition that an attribution map is
  better, if it changes the classification with less pixels changed.

To find a minimal binary mask $\mask_{\min}$, we search for a threshold of the
attribution map $\am$, such that the mask score $\Delta f(\xh) = f(\xh)_i -
f(\xc)_i$ (i.e., the change in classification score) is maximized while the
size of the mask is minimized, i.e., $\mask_{\min} = \argmin{\mask} |\mask| -
\Delta f(\xh)$ (where we omitted the dependency of $\xh$ on $\mask$ for
brevity).
  In order to minimize artifacts in the copying process we also apply a
  morphological closing operation with a window size of 10 pixels followed by a
  Gaussian Blur with $\sigma = 11 px$. The final masks highlight the relevant
  class discriminators by showing the user the counterfactual features, the
  original features they are replaced with, and the corresponding mask score
  $\Delta f(\xh)$, indicating the quantitative effect of the replacement on the
  classifier. See~\figref{fig:res_qual} for example pairs and
  corresponding areas $\mask_{\min}$.

Furthermore, by applying a sequence of thresholds for the attribution map
$\am$, we derive an objective evaluation procedure for a given attribution map:
  For each hybrid image $\xh$ in the sequence of thresholds, we consider the
  change in classifier prediction relative to the size of the mask that has
  been used to create the hybrid. We accumulate the change in classifier
  prediction over all mask sizes to derive our proposed DAC score. This
  procedure is explained in detail in \figref{fig:method:evaluation} for a
  single pair of images. When reporting the DAC score for a particular
  attribution method, we average the single DAC scores over all images, and all
  distinct pairs of classes.

\section{Experiments}

\begin{figure}
  \centerline{\input{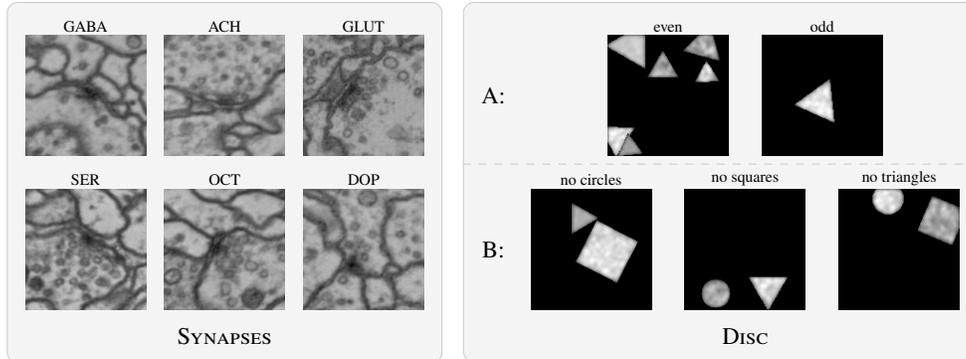}}
    \caption{Example images of datasets \synapse and \disc.
      \synapse consists of electron microscopy images of synapses. Each class
      is defined by the neurotransmitter the synapse releases. \disc is a
      synthetic dataset we designed in order to highlight failure cases of
      popular attribution methods. We consider two subsets: \disc-A shows
      triangles in each image and classes are defined by the parity of the
      number of triangles. \disc-B consists of images showing triangles,
      squares, and disks. Each class is one combination of two shapes.
  }
\label{fig:datasets}
\end{figure}

We evaluate the presented method on four datasets: \mnist~\citep{mnist},
\synapse~\citep{eckstein2020neurotransmitter}\footnote{Dataset kindly provided
by the authors of~\cite{eckstein2020neurotransmitter}.} and two versions of a
synthetic dataset that we call \disc-A and \disc-B (see~\figref{fig:datasets}
for an overview).

\paragraph{\synapse}
A real world biological dataset, consisting of $128\times 128$px electron
microscopy images of synaptic sites in the brain of \textit{Drosophila
melanogaster}.
  Each image is labelled with a functional property of the synapse, namely the
  neurotransmitter it releases (the label was acquired using
  immunohistochemistry labelling, see~\cite{eckstein2020neurotransmitter} for
  details).
  This dataset is of particular interest for interpretability, since a DNN can
  recover the neurotransmitter label from the images with high accuracy, but
  human experts are not able to do so. Interpretability methods like the one
  presented here can thus be used to gain insights into the relation between
  structure (from the electron microscopy image) and function (the
  neurotransmitter released).

\paragraph{\disc-A and \disc-B}
Two synthetic datasets with different discriminatory features of different
difficulty.
  Each image is $128\times 128$px in size and contains spheres, triangles or
  squares.
  For \disc-A, the goal is to correctly classify images containing an even or
  odd number of triangles.
  \disc-B contains images that show exactly two of the three available shapes
  and the goal is to predict which shape is missing (e.g., an image with only
  triangles and squares is to be classified as ``does not contain spheres'').
  This dataset was deliberately designed to investigate attribution methods in
  a setting where the discrimination depends on the absence of a feature.

\paragraph{Training}
For \mnist and \disc, we train a VGG and ResNet for 100 epochs and select the
epoch with highest accuracy on a held out validation dataset. For \synapse we
adapt the 3D-VGG architecture from~\cite{eckstein2020neurotransmitter} to 2D
and train for 500,000 iterations. We select the iteration with the highest
validation accuracy for testing. For each dataset we train one cycle-GAN for
200 epochs, on each class pair and on the same training set the respective
classifier was trained on (the full network specifications are given in the
supplement).

\begin{figure}
  \centerline{\input{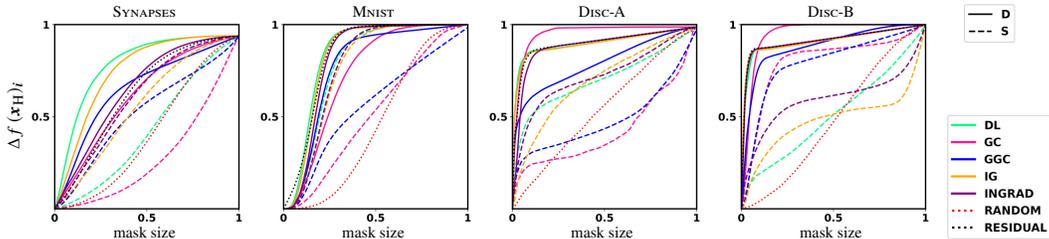}}
  \vspace{-1mm}
  \caption{Quantitative evaluation of discriminative ($D$ - solid) and
    corresponding original ($S$ for ``single input'' - dashed) attribution
    methods over four datasets.
    Corresponding $D$ and $S$ versions of the same method are shown in the same
    color. For each, we plot the average change of classifier prediction
    $\Delta f(\xh)^k_i = f(\xh)_i - f(\xc)_i$ as a function of mask size
    $\mask\in[0,1]$. In addition we show performance of the two considered
    baselines: masks derived from random attribution maps (random - red,
    dotted) and mask derived from the residual of the real and counterfactual
    image (residual - black, dotted). On all considered datasets all versions
    of $D$ attribution outperform their $S$ counterparts. All experiments 
    are performed with VGG architectures. For ResNet results of \mnist and \disc see supplement.}
  \label{fig:res_quant}
\end{figure}

\begin{figure}
\input{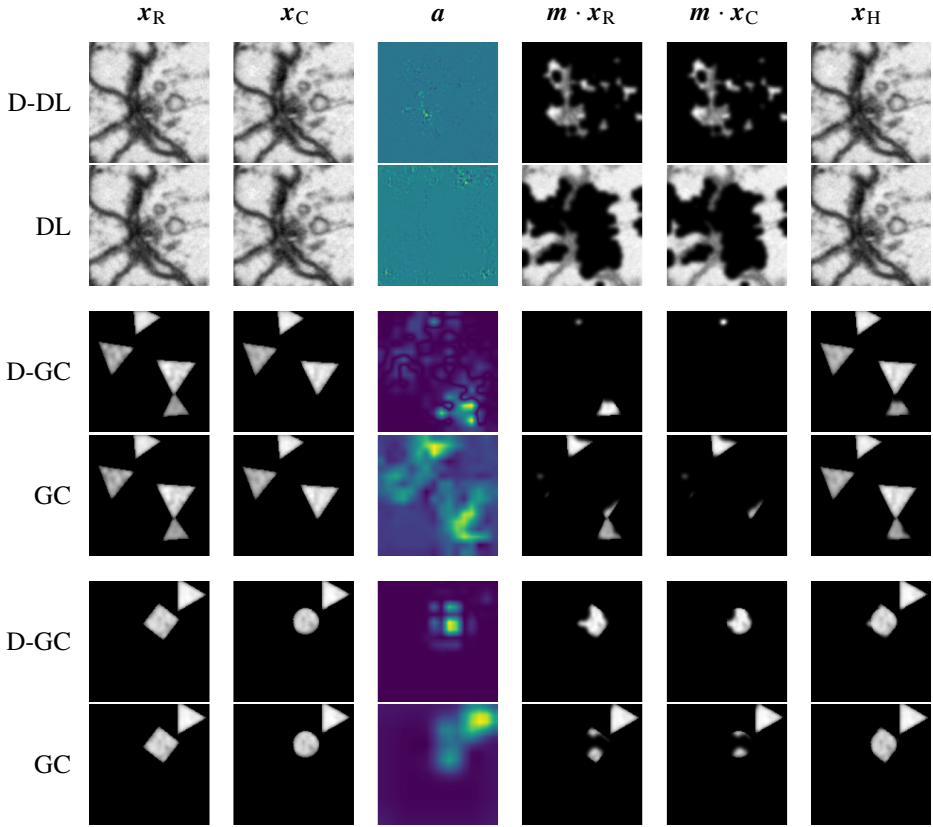}
  \caption{Samples from the best performing method pair on \synapse and
    \disc-A and B.
    Discriminative attribution methods are able to highlight the key
    discriminative features while vanilla versions often fail to do so (e.g., a
    subtle intensity change in the synaptic cleft in the top rows).
    Further qualitative results (including the other considered datasets) can
    be found in the supplement.
  }
\label{fig:res_qual}
\end{figure}

\paragraph{Results}
\label{sec:results}
Quantitative results (in terms of the DAC score, see
\secref{sec:method:evaluation}) for each investigated attribution method are
shown in \figref{fig:res_quant} and \tabref{tab:res_quant}.
  In summary, we find that attribution maps generated from the proposed
  discriminative attribution methods consistently outperform their original
  versions in terms of the DAC score.
  This observation also holds visually: the generated masks from discriminative
  attribution methods are smaller and more often highlight the main
  discriminatory parts of a considered image pair (see~\figref{fig:res_qual}).
  In particular, the proposed method substantially outperforms the considered
  random baseline, whereas standard attribution methods sometimes fail to do
  so (e.g., GC on dataset \synapse).
  Furthermore, on \mnist and \disc-A, the mask derived from the residual of real
  and counterfactual image is already competitive with the best considered
  methods and outperforms standard attribution substantially. However, for more
  complex datasets such as \synapse the residual becomes less accurate in
  highlighting discriminative features. Here, the discriminatory attributions
  outperform all other considered methods.

\begin{table}
  \centerline{\scale{0.9}{\footnotesize%
\pgfkeys{/pgf/number format/.cd,fixed,precision=2}%
\rowcolors{2}{gray!2!white}{gray!20!white}%
\pgfplotstableread[col sep=comma]{figures/data/results_v2/auc_vgg.csv}\data%
\pgfplotstabletypeset[%
  columns/Dataset/.style={string type},%
  every head row/.style={%
    before row={\toprule},%
    after row=\midrule},%
  every last row/.style={after row={\toprule}},%
  highlight row max={\data}{1},%
  highlight row maxx={\data}{2},%
  highlight row maxxx={\data}{3},%
  highlight row maxxxx={\data}{4},%
]{\data}%
}}
  \caption{Summary of DAC scores for each investigated method on the three
    datasets \mnist, \synapse, and \disc (two versions) corresponding to~\ref{fig:res_quant}.
    Best results are highlighted. For extended results with ResNet architectures see supplement.  
  }
  \label{tab:res_quant}
\end{table}

\section{Discussion}
\label{sec:discussion}

This work demonstrates that the combination of counterfactual interpretability
with suitable attribution methods is more accurate in extracting key
discriminative features between class pairs than standard methods.
  While the method succeeds in the presented experiments, it comes with a
  number of limitations.
  It requires the training of cycle-GANs, one for each pair of output classes.
  Thus training time and compute cost scale quadratically in the number of
  output classes and it is therefore not feasible for classification problems
  with a large number of classes. Furthermore, the translation from the real to
  the counterfactual image could fail for a large fraction of input images,
  i.e., $f(\xc)\neq j$. In this work, we only consider those image pairs where
  translation was successful, as we focus on extracting knowledge about class
  differences from the classifier. For applications that require an attribution
  for each input image this approach is not suitable.
  An additional concern is that focusing only on images that have a successful
  translation may bias the dataset we consider and with it the results. GANs
  are known to exhibit so called mode
  collapse~\citep{che2016mode,salimans2016improved}, meaning they focus on only
  a small set of modes of the full distribution. As a consequence, the method
  described here may miss discriminatory features present in other modes.
  Using a cycle-GAN is not possible in all image domains. Image classes need to
  be sufficiently similar in appearance for the cycle-GAN to work, and
  translating, e.g., an image of a mouse into an image of a tree is unlikely to
  work and produce meaningful attributions. However, we believe that the
  generation of masks in combination with the corresponding mask score is
  superior for interpreting DNN decision boundaries than classical attribution
  maps and suggest the usage of cycle-GAN baselines for attribution in cases
  where a fine grained understanding of class differences is sought.

Although we present this work in the context of understanding DNNs and the
features they make use of, an uncritical adaptation of this and other similar
interpretability methods can potentially lead to ethical concerns.
  As an example, results should be critically evaluated when using this method
  to interpret classifiers that have been trained to predict human behaviour,
  or demographic and socioeconomic features.
  As with any data-heavy method, it is important to realize that results will
  be reflective of data- and model-intrinsic biases. As such, an
  interpretability method like the one we present here can at most identify a
  correlation between input features and labels, but not true causal links.
  The method presented here should therefore not be used to ``proof'' that a
  particular feature leads to a particular outcome. Such claims should be met
  with criticism to prevent agenda-driven narratives of malicious actors.

{\small
\bibliographystyle{apalike}
\bibliography{references}
}

\clearpage

\appendix

\section{Training Details}

\subsection{Network Architectures}

\paragraph{Cycle-GAN}

We extend the cylce-GAN implementation from
\url{https://github.com/junyanz/pytorch-CycleGAN-and-pix2pix} for our purposes.
For all experiments we use a 9-block \resnet generator and a $70\times 70$
PatchGAN~\citep{isola2017image} discriminator. For training we use a least
squares loss (LSGAN~\citep{mao2017least}), a batch size of one, instance
normalization and normal initialization. We use the Adam
optimizer~\citep{kingma2014adam} with momentum $\beta_1=0.5$ and a learning
rate of $0.0002$ with a linear decay to zero after the first 100 epochs.

\paragraph{Classifiers}

\begin{table}[t]
  \rowcolors{2}{gray!2!white}{gray!20!white}
  \begin{subtable}[t]{0.4\textwidth}
    \begin{tabular}[t]{ll}
  \toprule
  Operation                             & Tensor Size \\
  \midrule

  input image                           & $(128, 128)$ \\

  \operation{Conv2d}, size $(3, 3)$     & $(12, 128, 128)$ \\
  \operation{BatchNorm2d}               & $(12, 128, 128)$ \\
  \operation{ReLU}                      & $(12, 128, 128)$ \\
  \operation{Conv2d}, size $(3, 3)$     & $(12, 128, 128)$ \\
  \operation{BatchNorm2d}               & $(12, 128, 128)$ \\
  \operation{ReLU}                      & $(12, 128, 128)$ \\
  \operation{MaxPool2d}, size $(2, 2)$  & $(12, 64, 64)$ \\

  \operation{Conv2d}, size $(3, 3)$     & $(24, 64, 64)$ \\
  \operation{BatchNorm2d}               & $(24, 64, 64)$ \\
  \operation{ReLU}                      & $(24, 64, 64)$ \\
  \operation{Conv2d}, size $(3, 3)$     & $(24, 64, 64)$ \\
  \operation{BatchNorm2d}               & $(24, 64, 64)$ \\
  \operation{ReLU}                      & $(24, 64, 64)$ \\
  \operation{MaxPool2d}, size $(2, 2)$  & $(24, 32, 32)$ \\

  \operation{Conv2d}, size $(3, 3)$     & $(48, 32, 32)$ \\
  \operation{BatchNorm2d}               & $(48, 32, 32)$ \\
  \operation{ReLU}                      & $(48, 32, 32)$ \\
  \operation{Conv2d}, size $(3, 3)$     & $(48, 32, 32)$ \\
  \operation{BatchNorm2d}               & $(48, 32, 32)$ \\
  \operation{ReLU}                      & $(48, 32, 32)$ \\
  \operation{MaxPool2d}, size $(2, 2)$  & $(48, 16, 16)$ \\

  \operation{Conv2d}, size $(3, 3)$     & $(96, 16, 16)$ \\
  \operation{BatchNorm2d}               & $(96, 16, 16)$ \\
  \operation{ReLU}                      & $(96, 16, 16)$ \\
  \operation{Conv2d}, size $(3, 3)$     & $(96, 16, 16)$ \\
  \operation{BatchNorm2d}               & $(96, 16, 16)$ \\
  \operation{ReLU}                      & $(96, 16, 16)$ \\
  \operation{MaxPool2d}, size $(2, 2)$  & $(96, 8, 8)$ \\

  \operation{Linear}                    & $(4096)$ \\
  \operation{ReLU}                      & $(4096)$ \\
  \operation{Dropout}                   & $(4096)$ \\
  \operation{Linear}                    & $(4096)$ \\
  \operation{ReLU}                      & $(4096)$ \\
  \operation{Dropout}                   & $(4096)$ \\

  \operation{Linear}                    & $(k)$ \\

  \toprule
\end{tabular}

    \caption{\vgg architecture used for the \synapse ($k=6$), \disc-A ($k=2$),
    and \disc-B ($k=3$) dataset.}
  \end{subtable}
  \hspace{\fill}
  \begin{subtable}[t]{0.4\textwidth}
    \begin{tabular}[t]{ll}
  \toprule
  Operation                             & Tensor Size \\
  \midrule

  input image                           & $(28, 28)$ \\

  \operation{Conv2d}, size $(3, 3)$     & $(12, 28, 28)$ \\
  \operation{BatchNorm2d}               & $(12, 28, 28)$ \\
  \operation{ReLU}                      & $(12, 28, 28)$ \\
  \operation{Conv2d}, size $(3, 3)$     & $(12, 28, 28)$ \\
  \operation{BatchNorm2d}               & $(12, 28, 28)$ \\
  \operation{ReLU}                      & $(12, 28, 28)$ \\
  \operation{MaxPool2d}, size $(2, 2)$  & $(12, 14, 14)$ \\

  \operation{Conv2d}, size $(3, 3)$     & $(24, 14, 14)$ \\
  \operation{BatchNorm2d}               & $(24, 14, 14)$ \\
  \operation{ReLU}                      & $(24, 14, 14)$ \\
  \operation{Conv2d}, size $(3, 3)$     & $(24, 14, 14)$ \\
  \operation{BatchNorm2d}               & $(24, 14, 14)$ \\
  \operation{ReLU}                      & $(24, 14, 14)$ \\
  \operation{MaxPool2d}, size $(2, 2)$  & $(24, 7, 7)$ \\

  \operation{Conv2d}, size $(3, 3)$     & $(48, 7, 7)$ \\
  \operation{BatchNorm2d}               & $(48, 7, 7)$ \\
  \operation{ReLU}                      & $(48, 7, 7)$ \\
  \operation{Conv2d}, size $(3, 3)$     & $(48, 7, 7)$ \\
  \operation{BatchNorm2d}               & $(48, 7, 7)$ \\
  \operation{ReLU}                      & $(48, 7, 7)$ \\

  \operation{Conv2d}, size $(3, 3)$     & $(96, 7, 7)$ \\
  \operation{BatchNorm2d}               & $(96, 7, 7)$ \\
  \operation{ReLU}                      & $(96, 7, 7)$ \\
  \operation{Conv2d}, size $(3, 3)$     & $(96, 7, 7)$ \\
  \operation{BatchNorm2d}               & $(96, 7, 7)$ \\
  \operation{ReLU}                      & $(96, 7, 7)$ \\

  \operation{Linear}                    & $(4096)$ \\
  \operation{ReLU}                      & $(4096)$ \\
  \operation{Dropout}                   & $(4096)$ \\
  \operation{Linear}                    & $(4096)$ \\
  \operation{ReLU}                      & $(4096)$ \\
  \operation{Dropout}                   & $(4096)$ \\

  \operation{Linear}                    & $(10)$ \\

  \toprule
\end{tabular}

    \caption{\vgg architecture used for the \mnist dataset.}
  \end{subtable}
  \caption{\vgg classifier network architectures.}
  \label{tab:supp:vggs}
\end{table}

\begin{table}[t]
  \rowcolors{2}{gray!2!white}{gray!20!white}
  \begin{subtable}[t]{0.4\textwidth}
    \begin{tabular}[t]{ll}
  \toprule
  Operation                             & Tensor Size \\
  \midrule

  input image                           & $(128, 128)$ \\

  \operation{Conv2d}, size $(3, 3)$     & $(12, 128, 128)$ \\
  \operation{BatchNorm2d}               & $(12, 128, 128)$ \\
  \operation{ReLU}                      & $(12, 128, 128)$ \\

  \operation{ResBlock}, stride $(2, 2)$ & $(12, 64, 64)$ \\
  \operation{ResBlock}                  & $(12, 64, 64)$ \\

  \operation{ResBlock}, stride $(2, 2)$ & $(24, 32, 32)$ \\
  \operation{ResBlock}                  & $(24, 32, 32)$ \\

  \operation{ResBlock}, stride $(2, 2)$ & $(48, 16, 16)$ \\
  \operation{ResBlock}                  & $(48, 16, 16)$ \\

  \operation{ResBlock}, stride $(2, 2)$ & $(96, 8, 8)$ \\
  \operation{ResBlock}                  & $(96, 8, 8)$ \\

  \operation{Linear}                    & $(4096)$ \\
  \operation{ReLU}                      & $(4096)$ \\
  \operation{Dropout}                   & $(4096)$ \\
  \operation{Linear}                    & $(4096)$ \\
  \operation{ReLU}                      & $(4096)$ \\
  \operation{Dropout}                   & $(4096)$ \\

  \operation{Linear}                    & $(k)$ \\

  \toprule
\end{tabular}

    \caption{\resnet architecture used for the \disc-A ($k=2$) and \disc-B
    ($k=3$) dataset.}
  \end{subtable}
  \hspace{\fill}
  \rowcolors{2}{gray!2!white}{gray!20!white}
  \begin{subtable}[t]{0.4\textwidth}
    \begin{tabular}[t]{ll}
  \toprule
  Operation                             & Tensor Size \\
  \midrule

  input image                           & $(28, 28)$ \\

  \operation{Conv2d}, size $(3, 3)$     & $(12, 28, 28)$ \\
  \operation{BatchNorm2d}               & $(12, 28, 28)$ \\
  \operation{ReLU}                      & $(12, 28, 28)$ \\

  \operation{ResBlock}, stride $(2, 2)$ & $(12, 14, 14)$ \\
  \operation{ResBlock}                  & $(12, 14, 14)$ \\

  \operation{ResBlock}, stride $(2, 2)$ & $(24, 7, 7)$ \\
  \operation{ResBlock}                  & $(24, 7, 7)$ \\

  \operation{ResBlock}, stride $(2, 2)$ & $(48, 3, 3)$ \\
  \operation{ResBlock}                  & $(48, 3, 3)$ \\

  \operation{ResBlock}, stride $(2, 2)$ & $(96, 1, 1)$ \\
  \operation{ResBlock}                  & $(96, 1, 1)$ \\

  \operation{Linear}                    & $(4096)$ \\
  \operation{ReLU}                      & $(4096)$ \\
  \operation{Dropout}                   & $(4096)$ \\
  \operation{Linear}                    & $(4096)$ \\
  \operation{ReLU}                      & $(4096)$ \\
  \operation{Dropout}                   & $(4096)$ \\

  \operation{Linear}                    & $(10)$ \\

  \toprule
\end{tabular}

    \caption{\resnet architecture used for the \mnist dataset.}
  \end{subtable}
  \caption{\resnet classifier network architectures.}
  \label{tab:supp:resnets}
\end{table}

The classifiers used for attribution are either \vgg (for datasets \synapse,
\mnist, and \disc) or \resnet (for datasets \mnist and \disc) architectures,
trained using a cross-entropy loss. Individual layers are shown
in~\tabref{tab:supp:vggs} and~\tabref{tab:supp:resnets}.

For the training of the \vgg network on the \synapse dataset, we use the same
strategy (including augmentations) as described
in~\cite{eckstein2020neurotransmitter}, with the only difference being that we
consider 2D images instead of 3D volumes. We did not attempt to train a \resnet
on the \synapse dataset.

For the training of the \vgg and \resnet architectures on the \mnist and \disc
datasets we did not make use of augmentations and trained each network for 100
epochs with a batch size of 32 using the Adam optimizer (learning rate
$10^{-4}$).

\subsection{Compute}

The most significant part of the compute costs come from training the
cycle-GANs. For each experiment, cycle-GAN training for 200 epochs took around
5 days on a single RTX 2080Ti GPU. For \mnist experiments we trained a total of
45 cycle GANs, 15 for \synapse, and 4 for \disc. In total this results in
roughly 320 GPU-days for cycle-GAN training. In contrast, attribution and mask
generation is comparatively cheap and takes between 1-3 hours on 20 RTX 2080Ti
GPUs for each dataset, resulting in 60 GPU hours for each experiment and 15 GPU
days in total.

\section{Extended Results for \resnet Architectures}

In addition to the results using \vgg architectures in the main text, below we
show additional results for \resnet architectures on \mnist and \disc-B (see
Fig.~\ref{fig:res_quant_sup} and Table~\ref{tab:res_quant_sup}).
  We do not show results for \disc-A, because all considered \resnet
  architectures failed to achieve more than chance level accuracy on the
  validation dataset. Since our goal is to understand what the classifier
  learned about class differences, using a network that did not successfully
  learn to classify will not produce meaningful results.

The shown results for \resnet architectures follow the same pattern as observed
in the main \vgg results:
  All discriminative attribution methods outperform their counterparts in terms
  of DAC-score. For \mnist, the overall best performing method is the residual,
  which already performed well for \vgg experiments. This is a consequence of
  the sparsity and simplicity of \mnist and can be observed less drastically
  for \disc as well. The changes the cycle-GAN introduces are often minimal,
  and thus the residual is already an accurate attribution. However, in
  general, the residual is not a good choice for an attribution map as
  intensity differences between classes do not generally correlate with feature
  importance. This is particularly noticeable in the experiments on the more
  challenging \synapse experiments (see main text).

\begin{figure}
\centerline{\input{figures/results_figures/eval_quant_res.tikz.tex}}
\caption{Quantitative evaluation of discriminative ($D$ - solid) and
    corresponding original ($S$ for ``single input'' - dashed) attribution
    methods for \mnist and \disc-B using a \textbf{\resnet} architecture. Corresponding $D$ and $S$ versions of the same method are shown in the same
    color. For each, we plot the average change of classifier prediction
    $\Delta f(\xh)^k_i = f(\xh)_i - f(\xc)_i$ as a function of mask size
    $\mask\in[0,1]$. In addition we show performance of the two considered
    baselines: masks derived from random attribution maps (random - red,
    dotted) and mask derived from the residual of the real and counterfactual
    image (residual - black, dotted). On all considered datasets all versions
    of $D$ attribution outperform their $S$ counterparts.}
\label{fig:res_quant_sup}
\end{figure}

\begin{table}
  \centerline{\scale{0.9}{\footnotesize%
\pgfkeys{/pgf/number format/.cd,fixed,precision=2}%
\rowcolors{2}{gray!2!white}{gray!20!white}%
\pgfplotstableread[col sep=comma]{figures/data/results_v2/auc_res.csv}\data%
\pgfplotstabletypeset[%
  columns/Dataset/.style={string type},%
  every head row/.style={%
    before row={\toprule},%
    after row=\midrule},%
  every last row/.style={after row={\toprule}},%
  highlight row max={\data}{1},%
  highlight row maxx={\data}{2},%
]{\data}%
}}
  \caption{Summary of DAC scores for \resnet architectures on \disc and \mnist corresponding to Fig.~\ref{fig:res_quant_sup}.
    Best results are highlighted.}
  \label{tab:res_quant_sup}
\end{table}

\begin{figure}
\centerline{\input{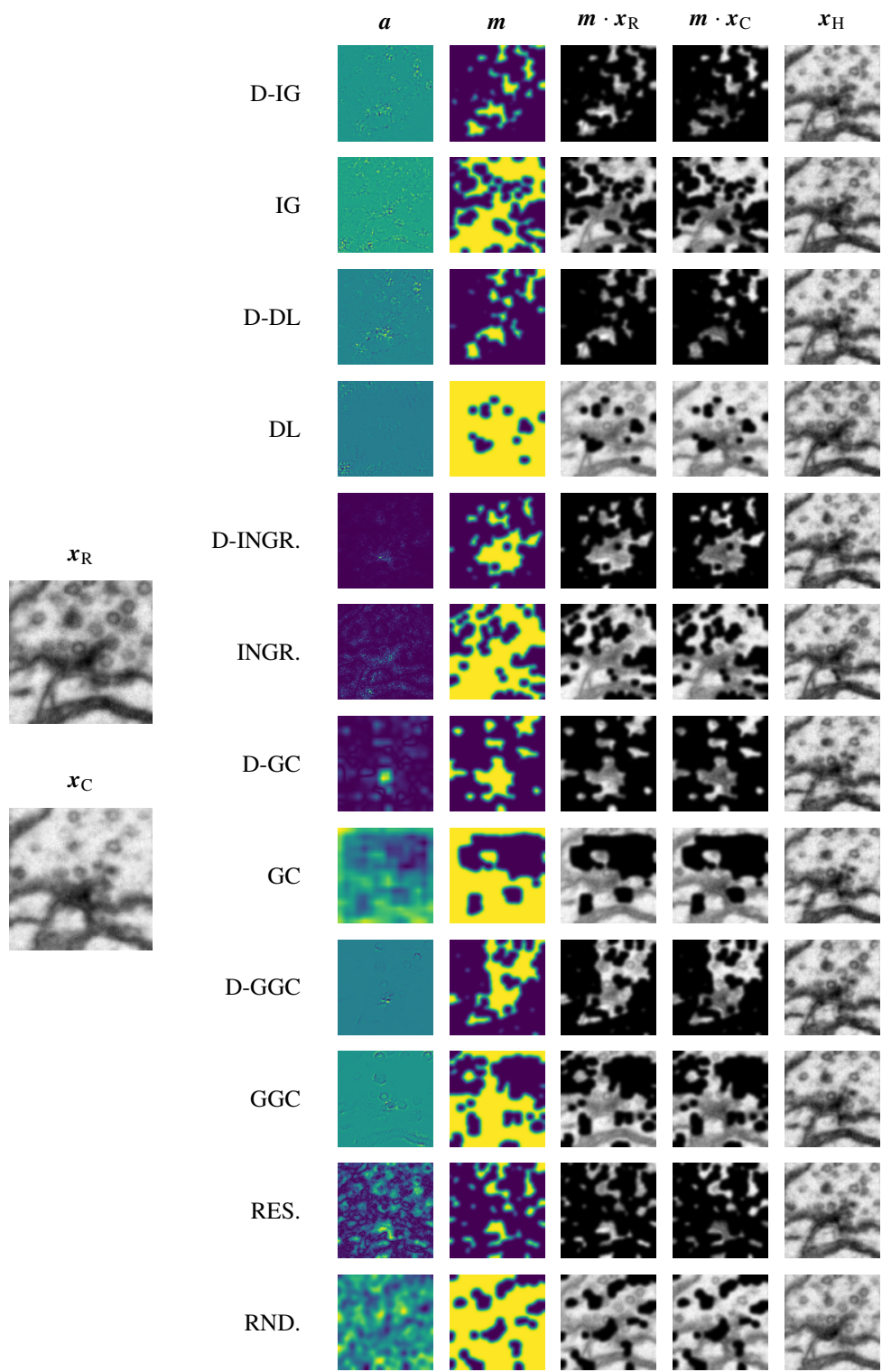}}
\caption{Qualitative samples from the \synapse dataset for all considered methods. $\xo$ shows a synapse from class Serotonin, $\xc$ shows a synapse from class Acetylcholine.}
\end{figure}

\begin{figure}
\centerline{\input{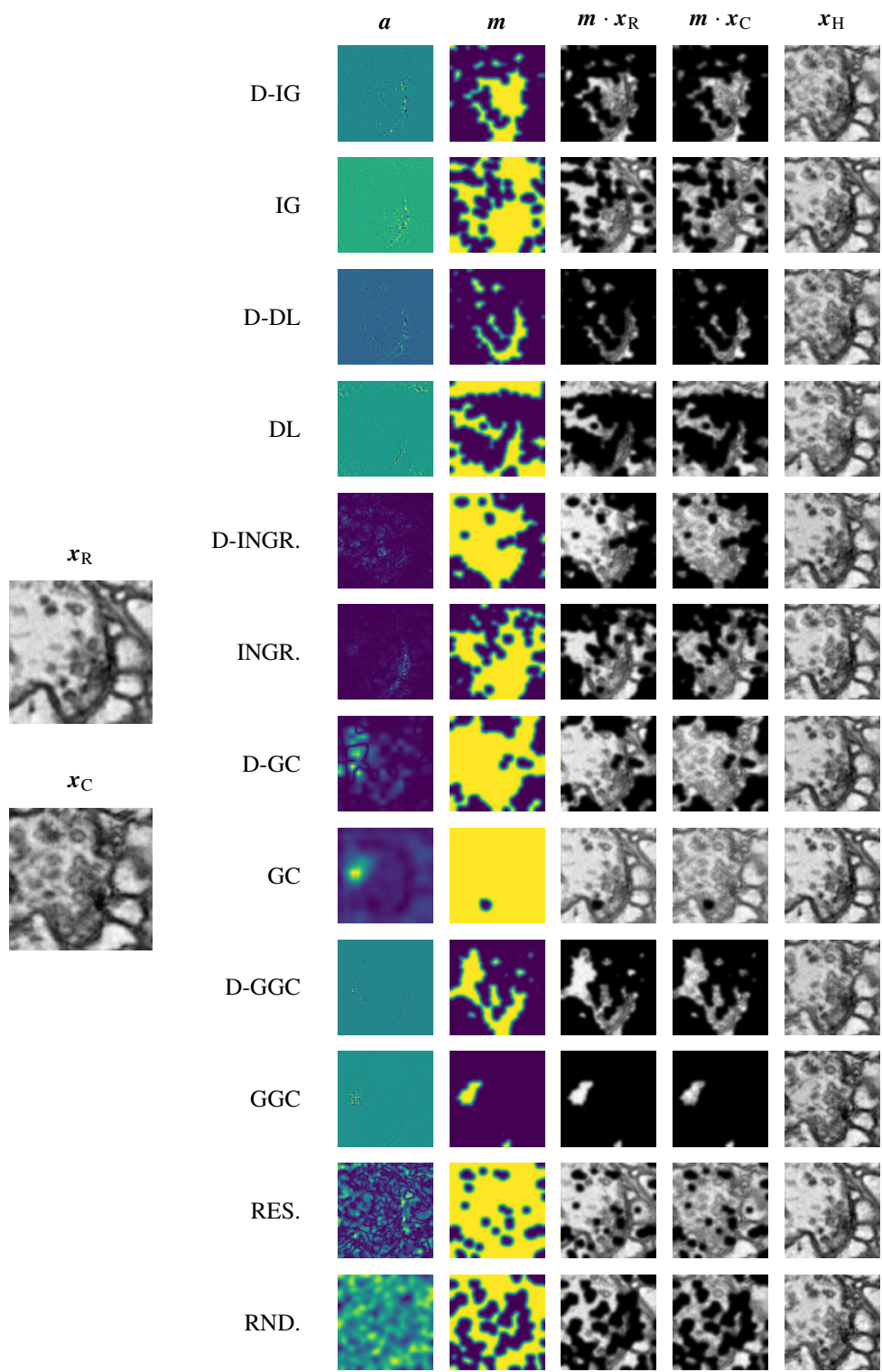}}
\caption{Qualitative samples from the \synapse dataset for all considered methods. $\xo$ shows a synapse from class Acetylcholine, $\xc$ shows a synapse from class Octopamine.}
\end{figure}

\begin{figure}
\centerline{\input{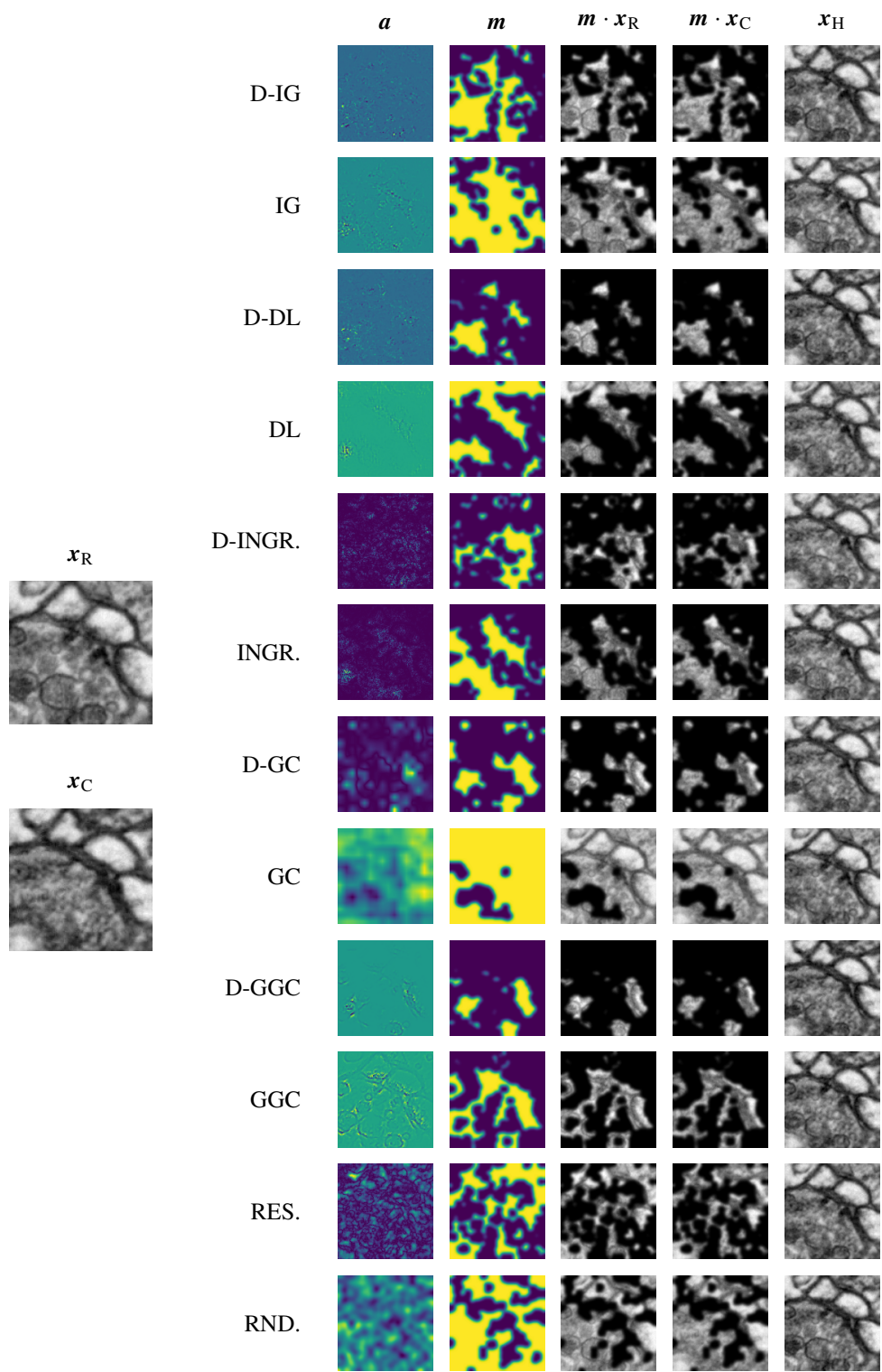}}
\caption{Qualitative samples from the \synapse dataset for all considered methods. $\xo$ shows a synapse from class Serotonin, $\xc$ shows a synapse from class Glutamate.}
\end{figure}

\begin{figure}
\centerline{\input{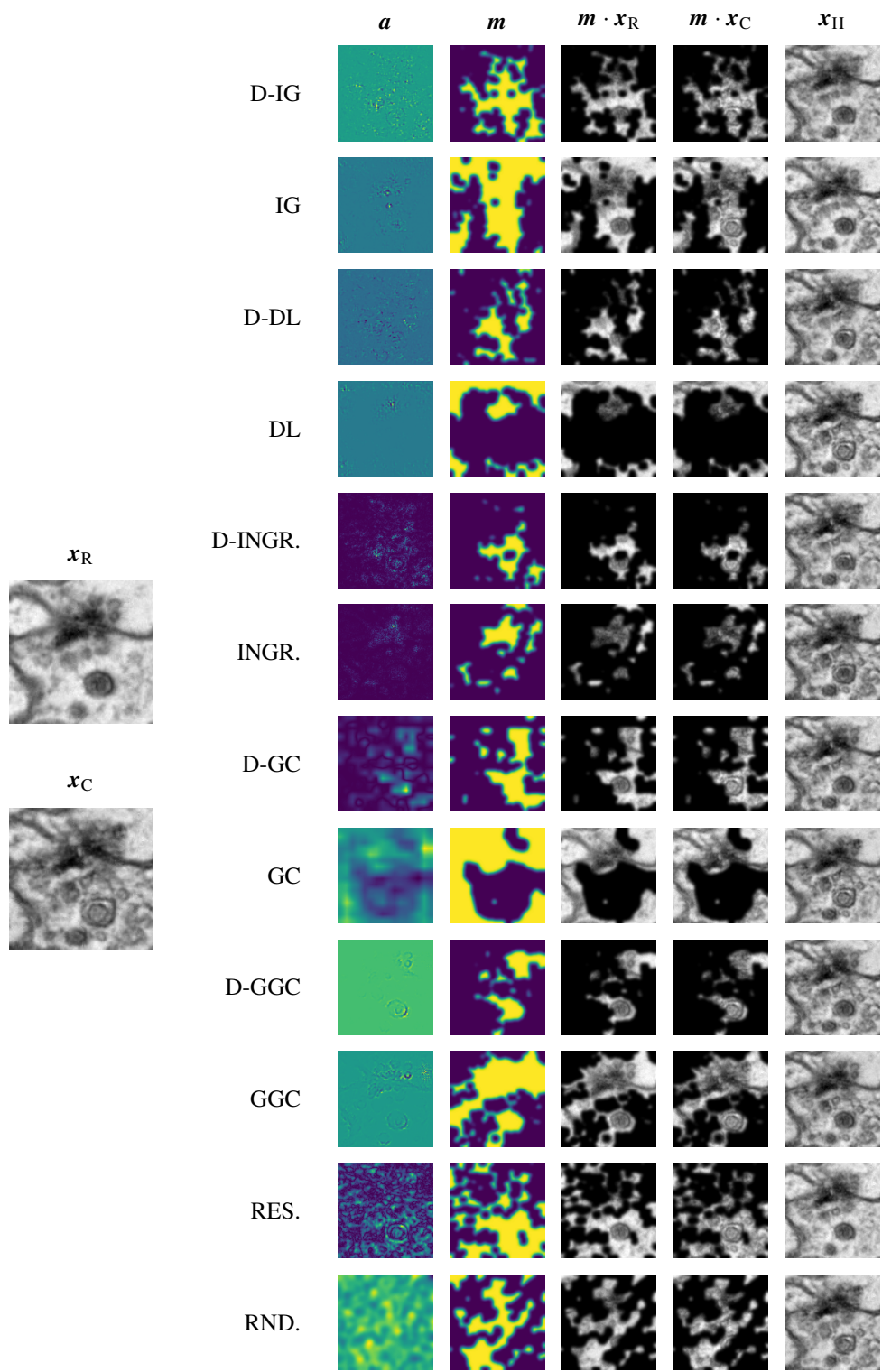}}
\caption{Qualitative samples from the \synapse dataset for all considered methods. $\xo$ shows a synapse from class Dopamine, $\xc$ shows a synapse from class Serotonin.}
\end{figure}

\begin{figure}
\centerline{\input{figures/results_figures/sup_qual/qual_sup_4.tikz.tex}}
\caption{Qualitative sample from the \mnist dateset for all considered methods.}
\end{figure}

\begin{figure}
\centerline{\input{figures/results_figures/sup_qual/qual_sup_6.tikz.tex}}
\caption{Qualitative sample from the \disc-A dateset for all considered methods.}
\end{figure}

\begin{figure}
\centerline{\input{figures/results_figures/sup_qual/qual_sup_7.tikz.tex}}
\caption{Qualitative sample from the \disc-B dateset for all considered methods.}
\end{figure}

\section{\disc Dataset}
The \disc dataset was specifically designed to highlight the advantage of discriminative attribution over vanilla attribution. In particular, the discriminatory feature of \disc-A is the parity of the number of triangles in the image. This feature is non-local and it is unclear what vanilla attribution is supposed to highlight. In \disc-B the classes are defined by the absence of a feature, another situation where vanilla attribution is not designed to give a sensible answer and will often highlight all objects in the image, providing little information to the user.\\
\paragraph{\disc-A} For each image we randomly draw an even (class 0) or odd (class 1) number between one and six, indicating the number of triangles to generate. Each triangle has a random size between 20 and 40\% of the image size of 128 pixels and a random position. In addition we draw a random intensity value between 120 and 200, a random rotation angle, and additive noise strength before applying Gaussian smoothing to generate different textures. We reject a sample if the fraction of foreground pixels and the total expected area of all shapes (assuming no overlap) is below 90\%, thus avoiding strongly overlapping configurations.

\paragraph{\disc-B} Similar to \disc-A, we draw a random position, intensity value, rotation and additive noise strength to generate images showing pairs of a triangle and a square, a disk and a square or a disk and a triangle. We reject a sample if the fraction of foreground pixels and the total expected area of all shapes (assuming no overlap) is below 90\%.

\section{Code and Data Availability}
All code, datasets, checkpoints, and instructions needed to reproduce the
presented results are available at \url{https://dac-method.github.io}.

\end{document}